\documentclass[letterpaper]{article} 
\usepackage{aaai25}  
\usepackage{times}  
\usepackage{helvet}  
\usepackage{courier}  
\usepackage[hyphens]{url}  
\usepackage{graphicx} 
\usepackage{natbib}  
\usepackage{caption} 
\usepackage{algorithm}
\usepackage{algorithmic}

\usepackage{newfloat}
\usepackage{listings}
\DeclareCaptionStyle{ruled}{labelfont=normalfont,labelsep=colon,strut=off} 
\floatstyle{ruled}
\newfloat{listing}{tb}{lst}{}
\floatname{listing}{Listing}
\usepackage{amsmath, amssymb}

\usepackage[table,xcdraw]{xcolor}

\title{Distribution-Free Uncertainty Quantification in Mechanical Ventilation Treatment: A Conformal Deep Q-Learning Framework}
\author{
     Niloufar Eghbali\textsuperscript{\rm 1},
    Tuka Alhanai\textsuperscript{\rm 2},
    Mohammad M. Ghassemi\textsuperscript{\rm1}
}
\affiliations{
        \textsuperscript{\rm 1}Michigan State University\\
    \textsuperscript{\rm 2}New York University \\
    eghbaliz@msu.edu, tuka.alhanai@nyu.edu, ghassem3@msu.edu\\
    \textsuperscript{\rm }Correspondance to Mohammad M. Ghassemi
    
}

\usepackage{bibentry}

\begin{document}

\maketitle

\begin{abstract}
Mechanical Ventilation (MV) is a critical life-support intervention in intensive care units (ICUs). However, optimal ventilator settings are challenging to determine because of the complexity of balancing patient-specific physiological needs with the risks of adverse outcomes that impact morbidity, mortality, and healthcare costs. This study introduces ConformalDQN, a novel distribution-free conformal deep Q-learning approach for optimizing mechanical ventilation in intensive care units. By integrating conformal prediction with deep reinforcement learning, our method provides reliable uncertainty quantification, addressing the challenges of Q-value overestimation and out-of-distribution actions in offline settings.
We trained and evaluated our model using ICU patient records from the MIMIC-IV database. ConformalDQN extends the Double DQN architecture with a conformal predictor and employs a composite loss function that balances Q-learning with well-calibrated probability estimation. This enables uncertainty-aware action selection, allowing the model to avoid potentially harmful actions in unfamiliar states and handle distribution shift by being more conservative in out-of-distribution scenarios.
Evaluation against baseline models, including physician policies, policy constraint methods, and behavior cloning, demonstrates that ConformalDQN consistently makes recommendations within clinically safe and relevant ranges, outperforming other methods by increasing the 90-day survival rate. Notably, our approach provides an interpretable measure of confidence in its decisions, crucial for clinical adoption and potential human-in-the-loop implementations.
\end{abstract}

%

\section{Introduction}

Mechanical ventilation (MV) is the most widely used short-term life support technique, frequently employed for planned surgical procedures, anesthesia, neonatal intensive care, life support during the
COVID-19 and acute organ failure \cite{coppola2014protective,mohlenkamp2020ventilation}. While mechanical ventilation is a life-saving intervention, it also carries the risk of significant complications \cite{pham2017mechanical}. Prolonged use, especially for durations of at least one week, can lead to substantial long-term effects on the physical, cognitive, and mental health of ICU survivors \cite{pham2017mechanical,zein2016ventilator}. Incorrect adjustments can trigger ventilator-induced lung injury (VILI), potentially leading to multiple organ dysfunction and significantly increasing mortality risk\cite{silva2022personalized}. Determining the most effective ventilator settings is challenging, as these settings must be tailored to each patient individually and it requires continuous re-evaluation of the optimal ventilation strategy during treatment \cite{zein2016ventilator}. In data-rich environments like intensive care units (ICUs), decision-making becomes more complicated and depends greatly on the skill and expertise of the attending physician \cite{slutsky2013ventilator}. The necessity for quick adjustments to ventilator settings in ICU can further elevate the risk of sub-optimal decisions, thereby increasing the likelihood of adverse outcomes. \cite{silva2022personalized}. 


Recent research has explored reinforcement learning (RL) algorithms for adaptive ventilator control, aiming to optimize treatment strategies based on patient-specific conditions \cite{suo2021machine,yu2021reinforcement,coronato2020reinforcement}. These approaches include decision tree-based RL \cite{lee2024methodology}, offline Batch Constraint Q-learning (BCQ) \cite{jingkun2024ventilator}, and Conservative Q-Learning \cite{yuan2023conservative}. RL agents can potentially adjust ventilator parameters in response to time-varying pulmonary mechanics, offering advantages over static protocol-driven methods \cite{jingkun2024ventilator}.
However, a significant challenge in applying RL to ventilator control is the state-action distribution shift problem. This occurs when the distribution of states and actions encountered during training differs from those in deployment, potentially leading to suboptimal or unsafe decisions in clinical settings. Current research has not adequately addressed this problem.

In this study, we focus on addressing the critical issue of distribution shift in RL-based mechanical ventilation control. We introduce a novel conformal deep Q-learning (ConformalDQN) framework that provides distribution-free uncertainty quantification. This approach aims to enhance the reliability and safety of RL-based decision-making in the dynamic and heterogeneous environment of ICU patients, without making strong distributional assumptions.
This framework allows the model to:\begin{itemize}
    \item Avoid potentially harmful actions in unfamiliar states.
    \item handle distribution shifts by being more conservative in out-of-distribution scenarios.
    \item Provide an interpretable measure of confidence in its decisions, which is crucial for clinical acceptance and potential human-in-the-loop implementations.
\end{itemize}
The ConformalDQN framework extends beyond mechanical ventilation treatment to other healthcare applications such as drug dosing and personalized treatment plans, where safety and adaptability to individual patient characteristics are essential. Moreover, its ability to handle uncertainty and distribution shifts makes it adaptable to safety-critical tasks beyond healthcare; it can be extended to autonomous driving and financial systems, where reliable decision-making under uncertainty is critical.

\section{Preliminaries}
This section provides essential background information and defines key concepts necessary for understanding our proposed ConformalDQN framework for mechanical ventilation optimization.
\subsection{Reinforcement Learning (RL)}
\begin{figure}[ht] 
    \centering
    \includegraphics[width=\columnwidth,height=0.7\columnwidth]{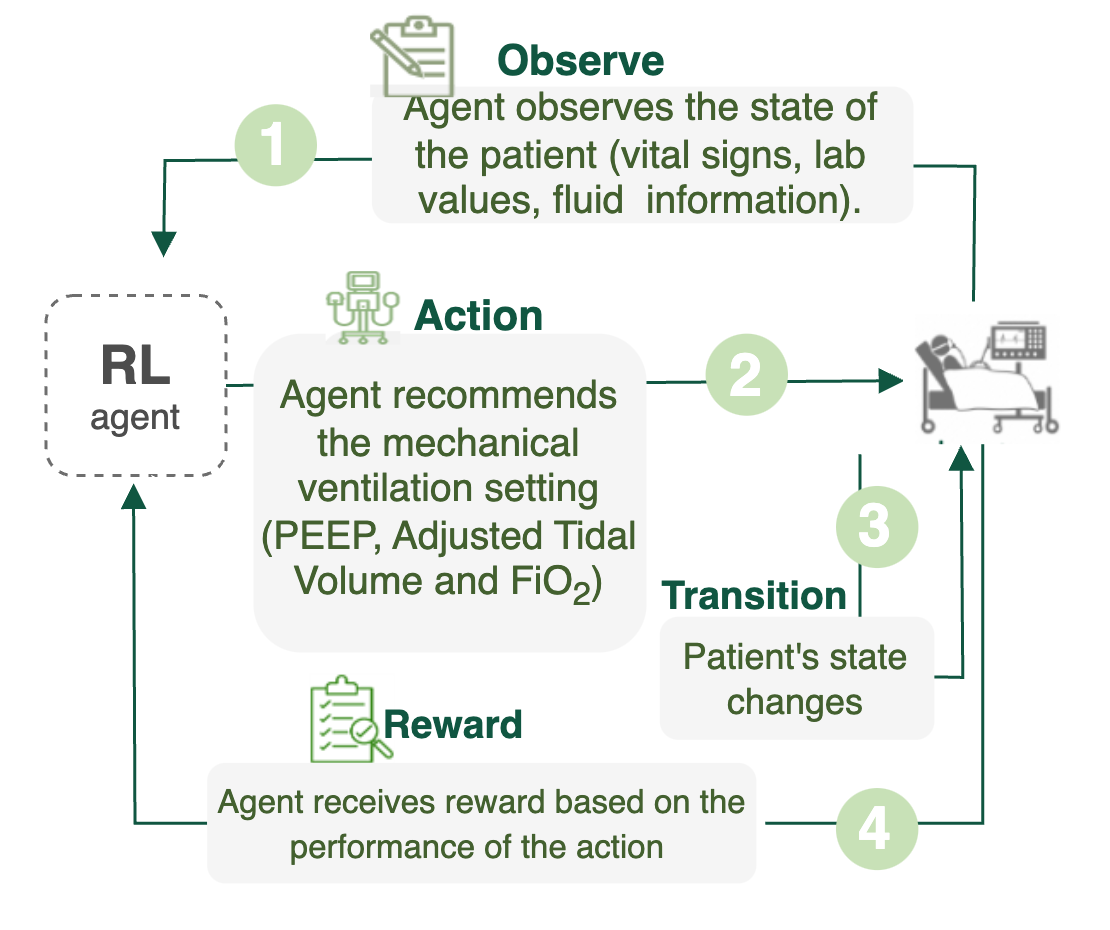} 
    \caption{ Interaction of RL agent with environment (patient).}
    \label{RL}
\end{figure}

We modeled the mechanical ventilation treatment problem as an RL problem formulated as a Markov Decision Process (MDP). This MDP is defined by the tuple $(S, A, P, R)$, where $s_t \in S$ represents the patient's state at time $t$, including physiological parameters; $a_t \in A$ denotes the action of adjusting ventilator parameters at time t; $P(s_{t+1}|s_t, a_t)$ is the transition probability to the next state given the current state and action; and $r(s_t, a_t) \in R$ is the reward function reflecting the effectiveness of ventilation settings. The RL agent aims to learn an optimal policy $\pi^*: S \rightarrow A$ that maximizes the expected cumulative discounted reward $J(\pi) = \mathbb{E}[\sum_{t=0}^T \gamma^t r(s_t, a_t)]$, where the discount factor $\gamma \in [0, 1]$ determines the relative importance of immediate rewards versus future rewards, with values closer to 1 emphasizing long-term consequences of actions. The agent iteratively observes the current state $s_t$, selects an action $a_t = \pi(s_t)$, receives a reward $r_t=r(s_t, a_t)$, and transitions to the next state $s_{t+1}$ (Figure \ref{RL}). Through this process, the agent learns to make decisions that optimize long-term patient outcomes while adapting to individual patient characteristics and changing conditions.

 \subsection{Bellman Equation}
A value function denoted as $Q(s,a)$ estimates the expected cumulative reward an agent can obtain by taking action $a$ in state $s$ and then following a certain policy.
The optimal Q-function, Q*(s,a), gives the maximum expected cumulative reward that can be obtained starting from state $s$ , taking action $a$ , and thereafter following the optimal policy $\pi^*$. Mathematically, it is defined by the Bellman optimality equation:
\begin{equation}
\fontsize{9pt}{8pt}
Q^*(s, a) = \sum_{s'\in S} P(s' \mid s, a) \left[ r(s, a) + \gamma \max_{a'\in A} Q^*(s', a') \right]
\end{equation}
The optimal policy $\pi^*$ selects the action with the highest $Q(s,a)$ in each state, maximizing the expected future rewards \cite{sutton1998reinforcement,mnih2015human}.

\subsection{Q-learning algorithms}
Q-learning is a model-free reinforcement learning algorithm that learns the optimal action-value function Q*(s,a) directly, without needing to learn the transition probabilities  \cite{watkins1992q}. The Q-learning update rule, derived from the Bellman error objective, is as follows:
\begin{equation}
\fontsize{9pt}{8pt}
Q(s_t,a_t) \leftarrow Q(s_t,a_t) + lr [r_t + \gamma \max_a Q(s_{t+1},a) - Q(s_t,a_t)]
\end{equation}
where $lr$ is the learning rate.

\noindent\underline{Deep Q-Network (DQN)} is a widely used reinforcement learning (RL) algorithm in healthcare applications \cite{yu2021reinforcement,fujimoto2019off}. It integrates Q-learning with deep neural networks to approximate the Q-function, denoted as $Q_{\theta}(s_t,a_t)$, using a neural network with parameters $\theta$. 

\noindent\underline{Double Deep Q-Network (DDQN)} is an enhanced version of DQN, designed for greater stability by utilizing two separate networks with identical architectures to estimate the target and prediction Q-values \cite{van2016deep}. The key concept involves using the prediction network, $Q_{\theta}$, for action selection, while the target network, $Q_{\theta'}$, is used for action evaluation. The target network’s output serves as the ground truth for updating the prediction network. The prediction network’s weights are updated with each iteration, whereas the target network’s weights are periodically updated by copying the prediction network’s weights. The goal is to minimize the loss function defined as:
\begin{equation}
\fontsize{8.5pt}{8pt}\selectfont
L(\theta) =   \left( r + \gamma  Q_{\theta'}(s_{t+1}, a'
) - Q_{\theta}(s_t, a_t) \right)^2,  a'= \arg\max_a Q_{\theta}(s, a) 
\end{equation}

\subsection{Offline Reinforcement Learning}

RL algorithms can be categorized into online and offline approaches. Online RL involves an agent learning through direct, real-time interaction with the environment, continuously updating its policy based on new experiences \cite{sutton1998reinforcement}. In contrast, Offline RL (also known as batch RL) learns from a fixed dataset of pre-collected experiences without active environment interaction \cite{levine2020offline,lange2012batch}. In Offline RL, the pre-collected dataset is typically generated by a behavioral policy, which is the policy used to collect the data; usually  a human expert. The goal of Offline RL is to learn an improved policy from this fixed dataset without further interaction with the environment. While online RL can adapt to changing environments and explore new strategies, it is often impractical and unsafe in critical domains like healthcare \cite{gottesman2019guidelines}. Therefore, Offline RL is more suitable for scenarios where live experimentation is risky or unethical  \cite{levine2020offline}. However, Offline RL faces unique challenges: 
\begin{enumerate}
\item Limited Data Coverage: The offline dataset may not cover all possible state-action pairs, leading to uncertainty in unexplored regions of the state-action space \cite{levine2020offline}. 
\item Distribution Shift: The distribution of states and actions in the offline dataset may differ from the distribution that would be encountered by the optimal policy \cite{fujimoto2019off,levine2020offline}. 
\item Extrapolation Error: Due to limited coverage, the learned value function may produce overconfident and overoptimistic estimates for out-of-distribution state-action pairs \cite{kumar2019stabilizing}. 
\end{enumerate}

\subsection{Policy constraints algorithms}
Policy constraint algorithms  aim to address the challenges of distribution shift and extrapolation error by explicitly constraining the learned policy to stay close to the behavior policy that generated the offline dataset \cite{kumar2020conservative,levine2020offline,fujimoto2019off}. 

One prominent approach in this category is Conservative Q-Learning (CQL) \cite{kumar2020conservative}. CQL learns a Q-function such that the expected policy value under this learned Q-function forms a lower bound on the actual policy value \cite{kumar2020conservative}. CQL adds a regularization term ($R_{CQL}$) to the standard Bellman error objective:
\begin{equation}
\fontsize{9pt}{6pt}\selectfont
R_{CQL}=\omega \mathbb{E}_{s \sim \mathcal{D}} [\log \sum_a \exp(Q(s,a)) - \mathbb{E}_{a \sim \pi_\beta(a|s)}[Q(s,a)]] ]
\end{equation}
where $\mathcal{D}$ is the offline dataset, $\pi_\beta$ is the behavior policy and $\omega$ is a hyperparameter controlling the degree of conservatism.
CQL  regularization term  encourages Q-values to be lower in expectation, counteracting the tendency to overestimate values for unseen state-action pairs. 

\subsection{Conformal Prediction }
Conformal prediction is a statistical framework for constructing prediction intervals or sets with guaranteed coverage probabilities under minimal distributional assumptions \cite{shafer2008tutorial}. Unlike traditional uncertainty quantification methods that often rely on specific probabilistic models, conformal prediction provides valid uncertainty estimates for any predictive model, including complex machine learning algorithms \cite{shafer2008tutorial,angelopoulos2021gentle}. The core idea involves using past observations to determine how 'unusual' a new prediction is, and then using this information to construct prediction sets. Formally, for a new input $x$ and a significance level $\alpha \in (0,1)$, conformal prediction constructs a prediction set $C(x)$ such that:
\begin{equation}
    \mathbb{P}(Y \in C(X)) \geq 1 - \alpha
\end{equation}
where $(X,Y)$ is a new test point. By calibrating these sets on a held-out calibration dataset, conformal prediction ensures that the true outcome falls within the predicted set with a specified probability $(1- \alpha)$, regardless of the underlying data distribution \cite{angelopoulos2021gentle}. The calibration process  involves computing nonconformity scores $s_i = s(x_i, y_i)$ for calibration points and setting a threshold $\tau$ such that:
\begin{equation} 
\fontsize{9pt}{6.5pt}
\tau = \text{Quantile}\left({s_i}_{i=1}^n, \left\lceil (n+1)(1-\alpha) \right\rceil / n\right) ]
\end{equation}
where $n$ is the number of calibration points. One commonly used nonconformity score for classification tasks is based on the softmax output of the predicted class probabilities. This score is defined as:
$s_i = s(x_i, y_i) = 1 - P(y_i|x_i)$
where $P(y_i|x_i)$ is the softmax probability of class $y_i$ given input $x_i$. This choice of nonconformity score reflects the model's confidence in its prediction, with lower scores indicating higher confidence.
For a new test sample x, the prediction set $C(x_i)$ is defined as:
\begin{equation} 
C(x_i) = \{y : s(x_i,y) \leq \tau\}
\end{equation} 
The flexibility and theoretical guarantees of conformal prediction have led to its application in various fields, including medical diagnosis where reliable uncertainty quantification is crucial for decision-making under risk \cite{balasubramanian2014conformal,vazquez2022conformal}.

\section{Related Works}
Recent advancements in reinforcement learning (RL) for mechanical ventilation optimization are highlighted by three key studies. Peine et al. (2021) introduced VentAI, pioneering tabular Q-learning with a 44-feature patient representation, optimizing key ventilation parameters and demonstrating potential to outperform clinical standards \cite{peine2021development}. Kondrup et al. (2022) developed DeepVent, an offline Deep RL agent using Conservative Q-Learning, expanding the state space and implementing a clinically relevant intermediate reward system to address sparse rewards and value overestimation \cite{kondrup2023towards}. The most recent work, DTE-CQL (2023), combines diagnosis and treatment aspects, integrating a Diagnose Transformer-Encoder for predicting patient states with CQL for treatment strategies, showing significant improvement over physician performance \cite{yuan2023conservative}. Despite these advancements, existing approaches struggle with uncertainty quantification, particularly in out-of-distribution scenarios common in heterogeneous patient populations. They also face challenges in providing interpretable and safe decision-making processes, essential for clinical adoption. Furthermore, the problem of distribution shift between offline training data and real-world deployment remains largely unaddressed, potentially leading to unreliable or unsafe decisions in practice.
To address these limitations, we introduce ConformalDQN, a novel approach that integrates conformal prediction with deep Q-learning. This method aims to provide distribution-free uncertainty quantification, enabling more reliable and safer decision-making in mechanical ventilation treatment. 

\begin{figure*}[ht] 
    \centering
    \includegraphics[width=1.8\columnwidth,height=5.7cm]{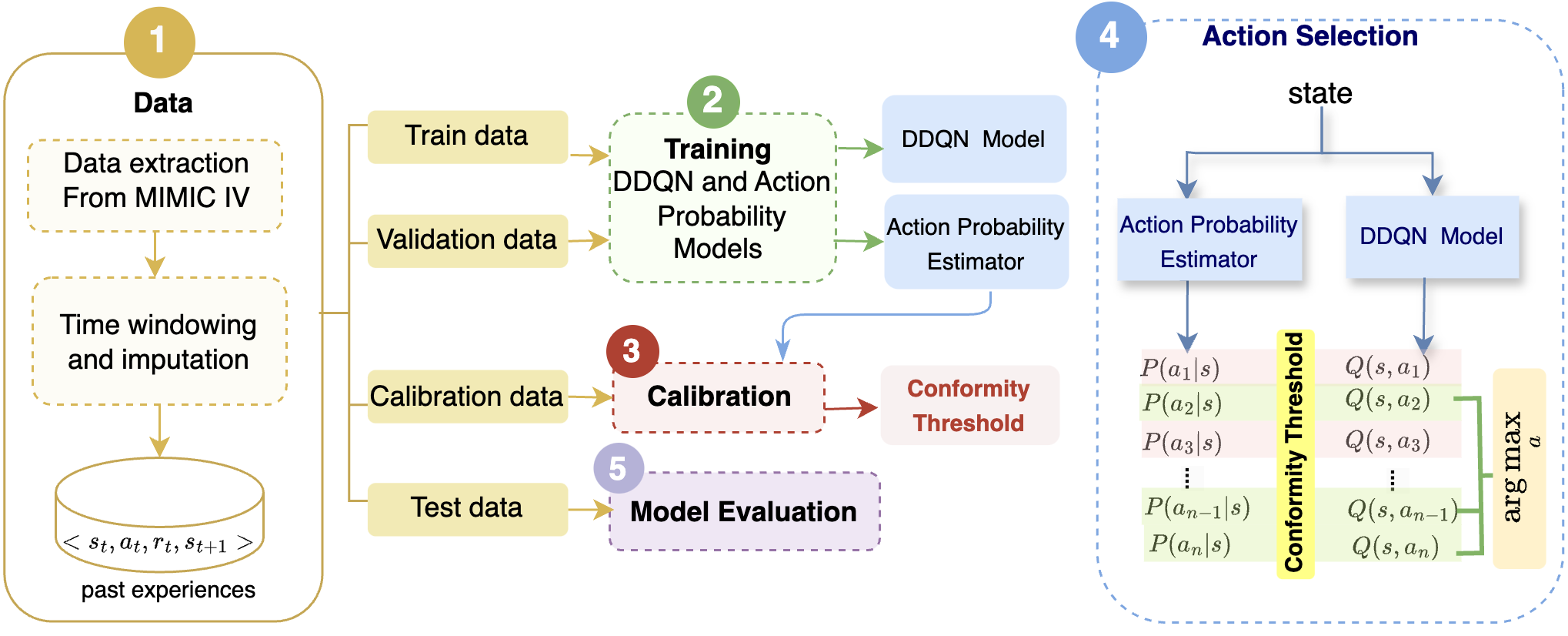} 
    \caption{Overview of the Method. (1)Preparing a structured dataset of past experiences.(2) Training a Double Deep Q-Network (DDQN) model and an action probability estimator (3) Calibration: utilizing the calibration set to determine a conformity threshold.(4) Action selection process: the state is fed into both the DDQN model and the action probability estimator, where actions are selected based on Q-values filtered through the conformity threshold to filter reliable actions. (5) Evaluation: the model's performance is evaluated using the test data. }
    \label{overview-fig}
\end{figure*}

\section{ Materials and Methods}
This section presents our proposed Conformal Deep Q-Learning (ConformalDQN) framework for optimizing mechanical treatment. 
\subsection{Data}
\subsubsection{Data Collection:} All data for this study were extracted from the Medical Information Mart for Intensive Care (MIMIC-IV) dataset \cite{johnson2020mimic}, a comprehensive, de-identified clinical database from an academic medical center in Boston, Massachusetts, USA. Using Standardized Query Language (SQL), we extracted 29,270 patients who underwent mechanical ventilation during their ICU stay.
\subsubsection{Data Preparation:} Following the work done by Kondrup et al. (2023), data preparation process involved the following three key steps:

\begin{enumerate}
    \item \noindent \underline{Temporal Segmentation:} We divided each patient's first 72 hours of ICU stay into 4-hour windows, resulting in 18 time windows per patient. 

   \item  \noindent\underline{Feature Collection:} For each window, we collected: vital signs, demographics, laboratory values, fluids  information, and ventilation settings (see appendix A for the glossary of acronyms ).
\begin{itemize}
    \item \textbf{Demographics:} age, gender, weight, ICU readmission, 90 days mortality, and elixhauser score.
    \item \textbf{Fluid Information:} urineoutput, intravenous fluids, cumulative fluid balance, vasopressors.
    \item \textbf{Vital Signs:} heart rate, sysBP, diasBP, meanBP, shock index, Respiratory rate, temperature, spo2, GCS, SOFA, SIRS.
    \item \textbf{Laboratory Values:} potassium, sodium, chloride, glucose, bun, creatinine, magnesium, calcium, ionizedcalcium, carbondioxide, bilirubin, hemoglobin, WBC, platelet, PTT, PT, INR, PH, PAO2, PACO, base excess, bicarbonate, loctate.
    \item \textbf{Ventilation settings:} Positive End-Expiratory Pressure (PEEP), weight-asjusted tidal volume (Vt) and FiO2.

\end{itemize}
\item \noindent\underline{Missing Data Handling:} 
We employed a hierarchical imputation strategy based on the extent of missing data: KNN for $<30\%$ missing, time-windowed sample-and-hold for 30-95\% missing, mean imputation for initial values, and removal for $>95\%$ missing data.
\end{enumerate}

\subsubsection{Data split:} We randomly divided patients into four sets: training
(60\%, 17,562 patients, 146,449 time windows), validation (10\%,
2,927 patients, 24,609 time windows), calibration (10\%,
2,927 patients, 23,605 time windows) and test set (20\%,
5,854 patients, 48,715 time windows). Training and validation sets were used for model development, calibration set for conformal prediction, and test set for final performance evaluation.

\subsubsection{Out-of-Distribution data selection:} Following Kondrup et al. (2023), we created an out-of-distribution (OOD) dataset for atypical cases. Patients were classified as OOD if any initial features fell within the top or bottom 1\% of the distribution at ventilation onset, resulting in approximately 20\% (5,942 patients, 61,014 time windows) classified as OOD.

\subsection{RL Problem Formulation}
We defined the MDP for mechanical ventilation based on Peine et al. (2021), with episodes from intubation to 72 hours post-initiation.\\
\textbf{State:} The state space encompasses 44 variables, including patient demographics (except 90-day mortality), vital signs, laboratory values, and fluid information. \\
\textbf{Action:} The action space is defined using a combination of three ventilator parameters: ideal weight-adjusted tidal volume (Vt), positive end-expiratory pressure (PEEP), and fraction of inspired oxygen (FiO2). Each of these parameters was discretized into seven treatment levels, representing specific ranges of settings. This resulted in a multi-dimensional action space of 343 discrete actions (7 x 7 x 7).\\
\textbf{Reward:} The reward function derived from the work of Kondrup et al. (2023); it combines a sparse terminal reward based on 90-day survival with an  intermediate reward based on changes in the patient's condition. The intermediate reward encourages the model to make decisions that improve the patient's immediate clinical status, while the terminal reward aligns the model's objectives with long-term patient outcomes.The reward function was formulated as:
\begin{equation}
    \fontsize{8pt}{8pt}
r(s_t, a_t, s_{t+1}) =
\begin{cases}
-1, \quad \text{   if $t=T$  and patient dies within 90 days} \\
+1,  \quad\text{   if $t=T$ and patient survives beyond 90 days} \\
\lambda \cdot \frac{\text{AP}(s_t) - \text{AP}(s_{t+1})}{\max_{AP} - \min_{AP}},  \quad\text{ otherwise}
\end{cases}
\end{equation}
 $AP$  is a modified version of the Acute Physiology and Chronic Health Evaluation II (APACHE II) score calculated for state $s$. $\max_{AP}$ and $\min_{AP}$ are the maximum and minimum possible values of the modified APACHE II score. APACHE II score is a critical care assessment tool that predicts the risk of mortality in ICU patients by evaluating physiological variables, age, and chronic health conditions \cite{knaus1991apache}.
$\lambda$ is a weighting factor for the intermediate reward.

\subsection{Conformal DQN Model}
 Conformal DQN  builds upon the standard DDQN architecture while incorporating a conformal predictor to provide distribution-free uncertainty estimates for decision-making. 
The key components of Conformal DQN include:
\begin{itemize}
    \item DDQN model that estimates the Q-values.
    \item A network $P_\omega$ that estimates the state conditional action probabilities of behavioral policy, $P_\omega(a|s)\approx \pi_{b}(a|s)$. 
    \item A conformal predictor that provides uncertainty estimates for each action; it determines how 'unusual' a new prediction is.
\end{itemize}

\subsubsection{Loss Function:} The Conformal DQN is trained using a composite loss function that combines multiple objectives: 
\begin{equation}
     L =  L_{\theta} + L_{NLL} + \lambda ||i||^2
\end{equation}
Where $L_{\theta}$ is the standard Double DQN loss describes as eq. 9 and 
$L_{NLL}$ is the negative log-likelihood loss of the selected actions based on the behavioral policy:
\begin{equation}
    L_{NLL} = -\log(P_\omega(a|s))]
\end{equation}
for the chosen action a, $||i||^2$ is the L2 norm of the raw logits. The loss function ensures the network learns Q-values for optimal action selection, encourages well-calibrated action probabilities for the conformal prediction mechanism, and includes L2 regularization on the raw logits to prevent overconfident predictions and improve generalization.

\subsubsection{Conformal predictor calibration:}
In this step, the nonconformity threshold will be calculated. This threshold represents a minimum level of confidence required for an action to be considered 'reliable'. We determine \(\tau\) using the following procedure: Let \(\{(s_i, a_i)\}_{i=1}^n\) be a calibration set of state-action pairs, and let \(1-\alpha\) be our desired confidence level. We compute \(\tau\) as follows:
\begin{equation}
   \tau = \text{Quantile}\left(\{1 - P_\omega(a_i|s_i)\}_{i=1}^n, \left\lceil (n+1)(1-\alpha) \right\rceil / n\right) 
\end{equation}

Where:
\begin{itemize}
    \item \(P_\omega(a_i|s_i)\) is the probability assigned by our model to action \(a_i\) in state \(s_i\)
    \item \(\text{Quantile}(S, q)\) returns the \(q\)-th quantile of the set \(S\)
    \item \(\alpha\) is the significance level (e.g., 0.05 for 95\% confidence)
\end{itemize}
The conformal predictor, calibrated on a held-out calibration dataset, provides a threshold \(\tau\) such that:
\begin{equation}
\fontsize{8pt}{8pt}
P(\text{} a^* \in \{a : P_\omega(a|s) \geq 1 - \tau\}) \geq 1 - \alpha
\end{equation}
where $a^*$ is an optimal behavioral action based on the dataset of experts' past experiences.

\subsubsection{Uncertainty-Aware Action Selection:}The action selection mechanism integrates the Q-values produced by the Double DQN with uncertainty estimates derived from the conformal predictor. The process of selecting the actions can be formalized as follows:

\begin{enumerate}
\item Creating the "confident' actions set: after selecting the threshold  $\tau$ a set of confident actions will be defined as: 
\begin{equation}
    A_c = \{a : P_{\omega}(a|s) \geq 1 - \tau\}
\end{equation}
\item Selecting the action: The action selection is then performed as follows:
\begin{equation}
a_t = \begin{cases}
\arg\max_{a \in A_c} Q(s,a) & \text{if } A_c \neq \emptyset \\
\arg\max_a Q(s,a) & \text{otherwise}
\end{cases}
\end{equation}
This mechanism ensures that we prioritize actions that are both high-value (according to the Q-function) and confident (according to the conformal predictor). More specifically, Actions with low probability (low confidence) are effectively filtered out. This means the model is avoiding actions it's uncertain about, even if they have high Q-values. In cases where no actions meet the confidence threshold, we employ the standard Q-learning action selection.
\end{enumerate}

\subsection{Evaluation}
We employed Fitted Q-Evaluation (FQE) \cite{le2019batch} to assess the performance of the trained policy. FQE provides more reliable estimates of a policy's value in offline reinforcement learning settings. It fits a Q-function to estimate the expected return using the Bellman equation. The FQE policy is trained on a sparse reward to assess the long-term outcome of 90-day survival.
\subsubsection{Baselines:} We evaluated ConformalDQN against several baselines: physician's policy, DeepVent(CQL) policy, and DDQN policy. 
\subsubsection{Evaluation metrics:} Performance was assessed using clinically relevant metrics including estimated 90-day survival rate, time within target physiological ranges, and frequency of unsafe ventilator settings.
\subsubsection{Out-of-Distribution (OOD) Evaluation:}
To assess the robustness of ConformalDQN in unfamiliar scenarios, we compared the performance of ConformalDQN against the baselines on both in-distribution (ID) and OOD data. We computed mean initial Q-values for each model in both settings. 


\subsection{Experimental Setups}
 We trained each model over five runs and averaged the performance. To determine the optimal set of hyperparameters, we conducted a grid search using uniform probability distribution across the following parameters: learning rates \([1e\text{-}3, 1e\text{-}4, 1e\text{-}5, 1e\text{-}6]\), discount factors \(\gamma\) \([0.05, 0.1, 0.5, 0.75, 0.9]\), number of hidden layers for the neural network \([1, 2, 3]\), and the size of units in each layer \([64, 128, 256, 512]\). 
For the \underline{CQL} model, the best results were achieved with a learning rate of \(1e\text{-}4\) and \(\omega = 0.1\). The \underline{ConformalDQN} model performed best with a learning rate of \(1e\text{-}3\) and \(\alpha = 0.15\). The \underline{DDQN} model performed best with a learning rate of \(1e\text{-}3\). The optimal architecture comprised two hidden layers with 256 units each, using the ReLU activation function. Models were trained with \(\gamma\)=0.75 for up to 30,000 timesteps, with the target Q-networks updated every 5,000 steps.

\section{Results}

To evaluate the model's performance, we conducted a correlation analysis between predicted Q-values and patient mortality. For each initial state-action pair (s, a) in the test dataset, we computed the model-predicted Q(s, a) and a binary mortality indicator M(s, a) (1 if the patient died, 0 otherwise). We then calculated the Pearson correlation coefficient  between the Q-values and mortality. As shown in Table \ref{tab:correlation_coefficients}, Conformal DQN demonstrates a stronger negative correlation between Q-values and mortality compared to CQL, indicating its superior performance in associating higher Q-values with lower mortality rates and predicting the patient outcomes. 
\begin{table}[ht]
    \centering
    \footnotesize
    \begin{tabular}{lcc}
        \textbf{Policy}& \textbf{Correlation Coefficient (Mean $\pm$ Std)}\\
        \hline
        \rowcolor[HTML]{EFEFEF} 
        ConformalDQN & -0.563 $\pm$ 0.003 & \\
        CQL & -0.429 $\pm$ 0.035 & \\
        \rowcolor[HTML]{EFEFEF} 
        DQN & -0.410 $\pm$ 0.011 & \\
        \hline
    \end{tabular}
    \caption{Comparison of correlation coefficients between predicted Q-values and patient mortality for different policies.}
    \label{tab:correlation_coefficients}
\end{table}

Table \ref{tab:performance} presents the mean Q-values of the initial patient states, as evaluated using Fitted Q-Evaluation (FQE) trained without intermediate rewards. These Q-values correlate with the expected 90-day survival rate, providing a measure of each policy's long-term effectiveness. We binned the physician's Q-values into intervals, calculated the observed 90-day survival rate for each bin, and mapped the mean Q-values of each policy to corresponding survival rates for interpretability.
\begin{table}[ht]
    \centering
     \footnotesize
    \begin{tabular}{lcc}
      & \textbf{Mean values of patient }  & \\
         \textbf{Policy}  & \textbf{initial state value}&\textbf{ Survival Rate(\%)}\\ 
        \hline
        \rowcolor[HTML]{EFEFEF} 
        Physician & 0.491 $\pm$ 0.001& 74.90\\
        DeepVent(CQL) & 0.581 $\pm$ 0.025 &81.65\\
        \rowcolor[HTML]{EFEFEF} 
        Conformal DQN & 0.639 $\pm$ 0.026&83.89 \\
        \hline
    \end{tabular}
\caption{Mean Q-values of initial patient states. Higher values indicate better expected long-term patient outcomes. }
    \label{tab:performance}
\end{table}
\noindent As shown in Table \ref{tab:performance}, ConformalDQN achieves the highest mean Q-value (0.639 ± 0.026), suggesting better performance in terms of expected 90-day survival rate. 
\subsection{Policy action distribution}
        
\begin{figure}[ht] 
    \centering
    \includegraphics[width=\columnwidth]{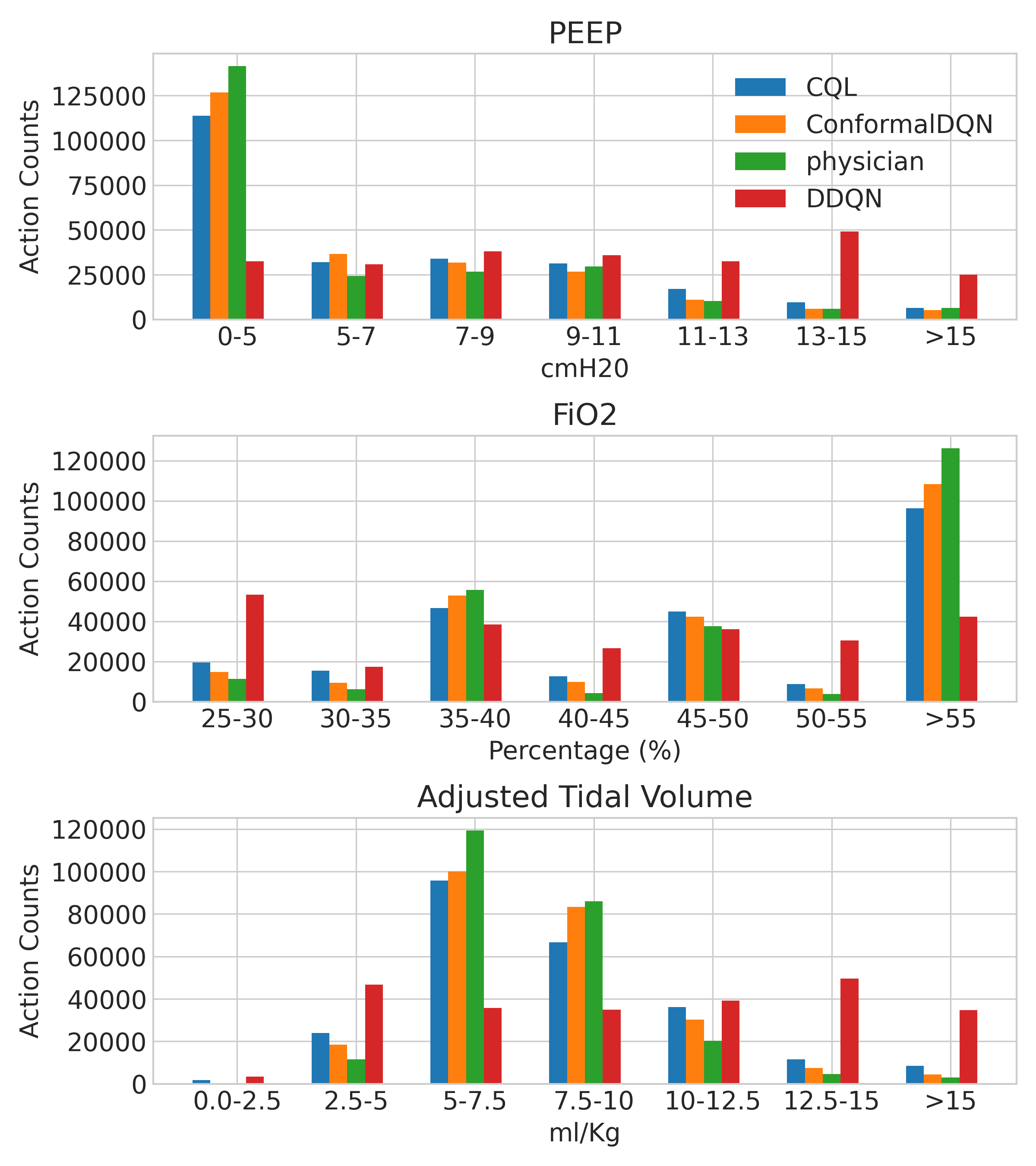} 
    \caption{ Distribution of ventilator settings chosen by different policies. PEEP: Positive End-Expiratory Pressure (cmH2O); FiO2: Fraction of Inspired Oxygen (\%); Adjusted Tidal Volume (ml/kg). }
    \label{actions}
\end{figure}

Figure \ref{actions} presents the action distributions for different mechanical ventilation policies, with a notable focus on ConformalDQN's clinically relevant choices. 
ConformalDQN demonstrates a preference for clinically conservative parameters: low PEEP (0-5 cmH2O), high FiO2 (\>55\%), and lung-protective tidal volumes (5-7.5 ml/kg). This distribution closely aligns with physician practices and established clinical guidelines, indicating ConformalDQN's capacity to learn and apply safe ventilation strategies that may reduce the risk of ventilator-induced lung injury.


\subsection{Out of distribution performance} 
\begin{figure}[ht] 
    \centering
    \includegraphics[width=\columnwidth,height=0.7\columnwidth]{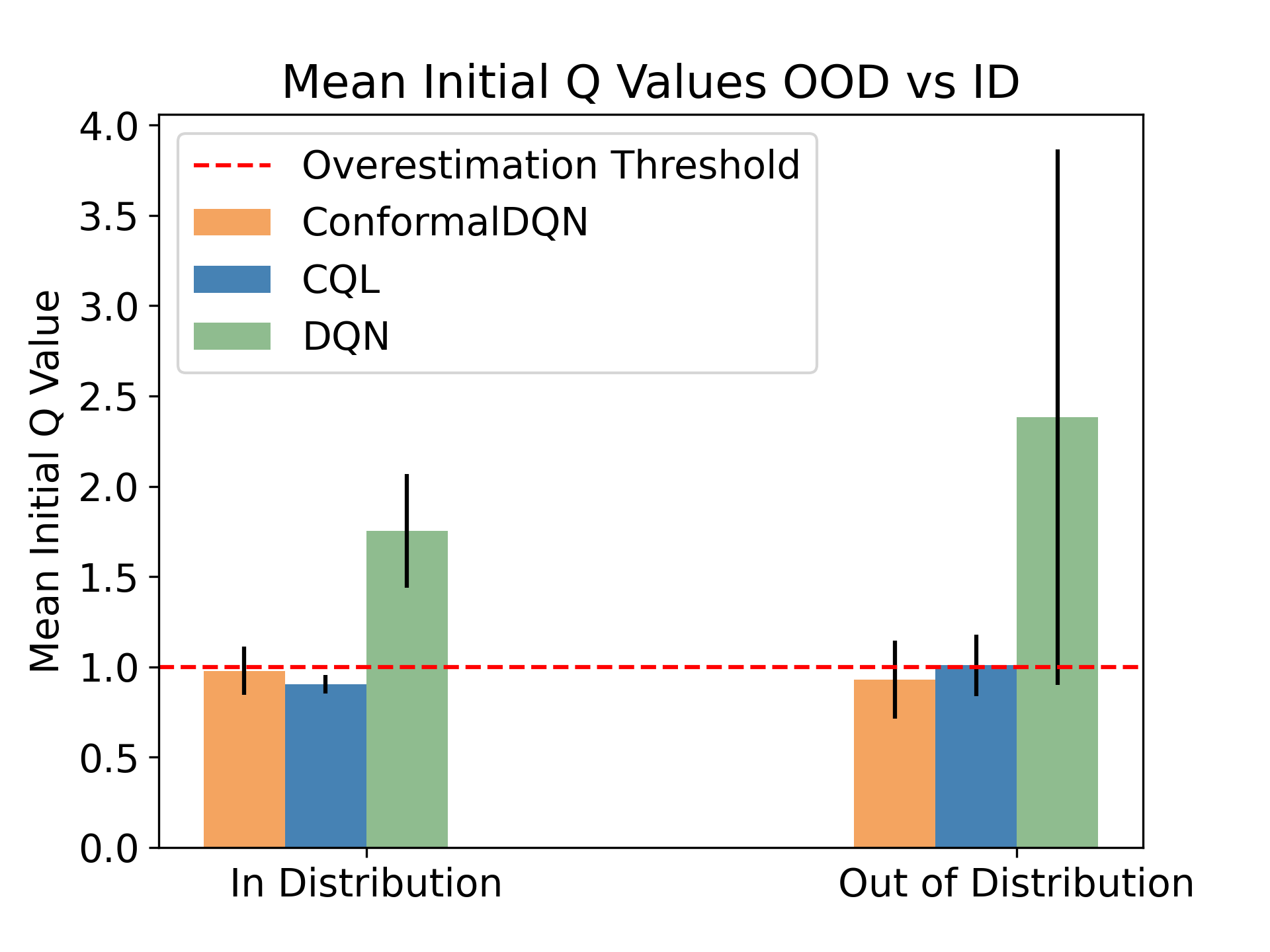} 
    \caption{ Comparison of mean initial Q-values for in-distribution (ID) and out-of-distribution (OOD) scenarios. Error bars represent variances. The horizontal line indicates the maximum expected return. }
    \label{ood}
\end{figure}

To assess the robustness of our models in unfamiliar scenarios, we conducted an out-of-distribution (OOD) evaluation. Figure \ref{ood}, performance of different models in both in-distribution (ID) and OOD data. The figure shows the mean initial Q-values, with a red dashed line indicating the overestimation threshold set at 1.0. This threshold represents the maximum possible return for an episode in the dataset, beyond which any Q-value is considered an overestimation.
The overestimation threshold is set at 1 because, the maximum return for an episode in the dataset—when no intermediate rewards are provided—is defined to be 1. This means that the best possible outcome, or the maximum expected cumulative reward that an agent can achieve by following an optimal policy, should not exceed 1.
Standard DDQN consistently overestimates Q-values, especially in OOD scenarios. This overestimation could lead to overly optimistic and potentially unsafe clinical decisions.
In contrast, CQL and ConformalDQN show stability across both in-distribution and OOD data, with Q-values remaining below the overestimation threshold. This stability validates ConformalDQN's approach to mitigating OOD state-action pair overestimation and demonstrates the effectiveness of ConformalDQN's uncertainty-aware method. These findings suggest that ConformalDQN's uncertainty-aware approach effectively mitigates the risks associated with OOD state-action pairs. By maintaining stable Q-value estimates and recommending more conservative actions in unfamiliar scenarios, ConformalDQN demonstrates enhanced robustness.

\section{Discussion}
In this study, we introduced ConformalDQN, a novel approach to mechanical ventilation management that integrates conformal prediction with deep Q-learning. Our results demonstrate the potential of this method to address critical challenges in applying reinforcement learning to healthcare, particularly the issues of distribution shift and uncertainty quantification in intensive care units.
ConformalDQN's performance, as evidenced by the mean Q-values of initial patient states (0.639 ± 0.026), indicates a significant improvement over CQL (0.581 ± 0.025) and physician policies (0.491 ± 0.001). This suggests that ConformalDQN could potentially lead to better long-term patient outcomes, specifically in terms of 90-day survival rates. The stronger negative correlation between Q-values and mortality compared to CQL and DQN further supports ConformalDQN's superior ability to assess patient risk and predict outcomes.

The action distribution analysis indicates that ConformalDQN closely mimics clinician behavior in selecting ventilator settings, especially in adhering to lung-protective strategies. The preference for lower PEEP values (0-5 cmH2O) and conservative tidal volumes (5-7.5 ml/kg) aligns with current clinical guidelines aimed at minimizing ventilator-induced lung injury. This alignment suggests that ConformalDQN has effectively learned clinically relevant patterns from the training data, a crucial factor for the potential clinical adoption of AI-driven decision support systems.

Importantly, ConformalDQN's performance in out-of-distribution scenarios demonstrates its robustness to the challenge of distribution shift. Unlike standard DDQN, which shows significant overestimation in OOD cases, ConformalDQN maintains stable Q-value estimates below the overestimation threshold. This characteristic is vital in healthcare applications, where encountering atypical cases is common and where overconfident decisions could lead to adverse outcomes.

The threshold in ConformalDQN acts as a tunable filter for action selection; by adjusting it to be less conservative (using a lower threshold value), we can allow the agent to consider actions that appear less frequently in the behavioral policy. This would enable more exploration since actions with lower probabilities in the expert dataset would still be considered valid choices if their Q-values are promising. A key advantage of ConformalDQN's approach is that adjusting the confidence threshold can be done post-training without requiring the model to be retrained. This flexibility allows us to tune the exploration-conservatism trade-off simply by changing the threshold parameter during deployment. The choice of conformal prediction over other uncertainty quantification techniques, such as Bayesian methods and ensembles, was driven by three key advantages: (1) it provides distribution-free uncertainty guarantees without requiring assumptions about physiological parameters and ventilation responses, (2) it is computationally efficient as a lightweight post-processing step, avoiding the overhead of ensemble models or posterior sampling, and (3) it allows clinicians to adjust conservatism thresholds after training without model retraining.

Compared to recent works in the field, such as VentAI (Peine et al., 2021) and DeepVent (Kondrup et al., 2023), ConformalDQN offers an additional advantage of uncertainty quantification. This feature not only enhances the safety of the model but also provides interpretable confidence measures, which could be crucial for clinician trust and potential human-in-the-loop implementations. 

However, despite the promising performance demonstrated by ConformalDQN in this study, further efforts need to be taken in order to be clinically applicable. An important direction for future research is the extension of ConformalDQN to continuous action spaces. The current discretized approach, while effective, may not capture the full granularity of ventilator settings possible in ICUs. By adapting the method to handle continuous actions, we could enable finer-grained control over ventilator parameters.  
Another direction is the development of a state-conditioned conformal prediction method. In this enhanced version, the conformal predictor would adapt its uncertainty estimates based on the current patient state, allowing for more nuanced and context-specific decision-making. This could potentially improve the model's performance in highly variable clinical scenarios, where the appropriate level of certainty may differ depending on the patient's condition.
\section{Conclusion}
In conclusion, ConformalDQN represents a promising step towards more reliable and safer AI-driven decision support in healthcare, particularly in mechanical ventilation management. By addressing key challenges in offline RL including uncertainty quantification and distribution shift, this approach offers  more robust and trustworthy AI systems. Notably, the application of this method can be extended beyond mechanical ventilation to other domains and safety-critical environments, such as autonomous driving and financial systems.

\section{Code Availability}
The code implementation of ConformalDQN is available on GitHub at https://github.com/HAAIL/ConformalDQN.

\section { Appendix A: Glossary of Acronyms}
This glossary provides the definitions for the acronyms used throughout the feature collection section.

\begin{itemize}
    \item \textbf{BUN:} Blood Urea Nitrogen
    \item \textbf{diasBP:} Diastolic Blood Pressure
    \item \textbf{FiO2:} Fraction of Inspired Oxygen
    \item \textbf{GCS:} Glasgow Coma Scale
    \item \textbf{INR:} International Normalized Ratio
    \item \textbf{PEEP:} Positive End-Expiratory Pressure
    \item \textbf{PaO2:} Partial Pressure of Oxygen
    \item \textbf{PaCO2:} Partial Pressure of Carbon Dioxide
    \item \textbf{PTT:} Partial Thromboplastin Time
    \item \textbf{SIRS:} Systemic Inflammatory Response Syndrome
    \item \textbf{SOFA:} Sequential Organ Failure Assessment
    \item \textbf{SpO2:} Oxygen Saturation
    \item \textbf{sysBP:} Systolic Blood Pressure
    \item \textbf{Vt:} Tidal Volume
    \item \textbf{WBC:} White Blood Cell count
\end{itemize}

\begingroup
\bibliography{aaai25}     

\begin{thebibliography}{31}
\providecommand{\natexlab}[1]{#1}

\bibitem[{Angelopoulos and Bates(2021)}]{angelopoulos2021gentle}
Angelopoulos, A.~N.; and Bates, S. 2021.
\newblock A gentle introduction to conformal prediction and distribution-free uncertainty quantification.
\newblock \emph{arXiv preprint arXiv:2107.07511}.

\bibitem[{Balasubramanian, Ho, and Vovk(2014)}]{balasubramanian2014conformal}
Balasubramanian, V.; Ho, S.-S.; and Vovk, V. 2014.
\newblock \emph{Conformal prediction for reliable machine learning: theory, adaptations and applications}.
\newblock Newnes.

\bibitem[{Coppola, Froio, and Chiumello(2014)}]{coppola2014protective}
Coppola, S.; Froio, S.; and Chiumello, D. 2014.
\newblock Protective lung ventilation during general anesthesia: is there any evidence?
\newblock \emph{Critical Care}, 18: 1--7.

\bibitem[{Coronato et~al.(2020)Coronato, Naeem, De~Pietro, and Paragliola}]{coronato2020reinforcement}
Coronato, A.; Naeem, M.; De~Pietro, G.; and Paragliola, G. 2020.
\newblock Reinforcement learning for intelligent healthcare applications: A survey.
\newblock \emph{Artificial intelligence in medicine}, 109: 101964.

\bibitem[{Fujimoto, Meger, and Precup(2019)}]{fujimoto2019off}
Fujimoto, S.; Meger, D.; and Precup, D. 2019.
\newblock Off-policy deep reinforcement learning without exploration.
\newblock In \emph{International conference on machine learning}, 2052--2062. PMLR.

\bibitem[{Gottesman et~al.(2019)Gottesman, Johansson, Komorowski, Faisal, Sontag, Doshi-Velez, and Celi}]{gottesman2019guidelines}
Gottesman, O.; Johansson, F.; Komorowski, M.; Faisal, A.; Sontag, D.; Doshi-Velez, F.; and Celi, L.~A. 2019.
\newblock Guidelines for reinforcement learning in healthcare.
\newblock \emph{Nature medicine}, 25(1): 16--18.

\bibitem[{Jingkun et~al.(2024)Jingkun, Fengxi, Chunxin, and Pixuan}]{jingkun2024ventilator}
Jingkun, M.; Fengxi, L.; Chunxin, L.; and Pixuan, Z. 2024.
\newblock Ventilator Treatment Policy Control based on BCQ off-line Deep Reinforcement Learning.

\bibitem[{Johnson et~al.(2020)Johnson, Bulgarelli, Pollard, Horng, Celi, and Mark}]{johnson2020mimic}
Johnson, A.; Bulgarelli, L.; Pollard, T.; Horng, S.; Celi, L.~A.; and Mark, R. 2020.
\newblock Mimic-iv.
\newblock \emph{PhysioNet. Available online at: https://physionet. org/content/mimiciv/1.0/(accessed August 23, 2021)}, 49--55.

\bibitem[{Knaus et~al.(1991)Knaus, Wagner, Draper, Zimmerman, Bergner, Bastos, Sirio, Murphy, Lotring, Damiano et~al.}]{knaus1991apache}
Knaus, W.~A.; Wagner, D.~P.; Draper, E.~A.; Zimmerman, J.~E.; Bergner, M.; Bastos, P.~G.; Sirio, C.~A.; Murphy, D.~J.; Lotring, T.; Damiano, A.; et~al. 1991.
\newblock The APACHE III prognostic system: risk prediction of hospital mortality for critically III hospitalized adults.
\newblock \emph{Chest}, 100(6): 1619--1636.

\bibitem[{Kondrup et~al.(2023)Kondrup, Jiralerspong, Lau, de~Lara, Shkrob, Tran, Precup, and Basu}]{kondrup2023towards}
Kondrup, F.; Jiralerspong, T.; Lau, E.; de~Lara, N.; Shkrob, J.; Tran, M.~D.; Precup, D.; and Basu, S. 2023.
\newblock Towards safe mechanical ventilation treatment using deep offline reinforcement learning.
\newblock In \emph{Proceedings of the AAAI Conference on Artificial Intelligence}, volume~37, 15696--15702.

\bibitem[{Kumar et~al.(2019)Kumar, Fu, Soh, Tucker, and Levine}]{kumar2019stabilizing}
Kumar, A.; Fu, J.; Soh, M.; Tucker, G.; and Levine, S. 2019.
\newblock Stabilizing off-policy q-learning via bootstrapping error reduction.
\newblock \emph{Advances in neural information processing systems}, 32.

\bibitem[{Kumar et~al.(2020)Kumar, Zhou, Tucker, and Levine}]{kumar2020conservative}
Kumar, A.; Zhou, A.; Tucker, G.; and Levine, S. 2020.
\newblock Conservative q-learning for offline reinforcement learning.
\newblock \emph{Advances in Neural Information Processing Systems}, 33: 1179--1191.

\bibitem[{Lange, Gabel, and Riedmiller(2012)}]{lange2012batch}
Lange, S.; Gabel, T.; and Riedmiller, M. 2012.
\newblock Batch reinforcement learning.
\newblock In \emph{Reinforcement learning: State-of-the-art}, 45--73. Springer.

\bibitem[{Le, Voloshin, and Yue(2019)}]{le2019batch}
Le, H.; Voloshin, C.; and Yue, Y. 2019.
\newblock Batch policy learning under constraints.
\newblock In \emph{International Conference on Machine Learning}, 3703--3712. PMLR.

\bibitem[{Lee, Mahendra, and Aswani(2024)}]{lee2024methodology}
Lee, J.~S.; Mahendra, M.; and Aswani, A. 2024.
\newblock Methodology for Interpretable Reinforcement Learning for Optimizing Mechanical Ventilation.
\newblock \emph{arXiv preprint arXiv:2404.03105}.

\bibitem[{Levine et~al.(2020)Levine, Kumar, Tucker, and Fu}]{levine2020offline}
Levine, S.; Kumar, A.; Tucker, G.; and Fu, J. 2020.
\newblock Offline reinforcement learning: Tutorial, review, and perspectives on open problems.
\newblock \emph{arXiv preprint arXiv:2005.01643}.

\bibitem[{Mnih et~al.(2015)Mnih, Kavukcuoglu, Silver, Rusu, Veness, Bellemare, Graves, Riedmiller, Fidjeland, Ostrovski et~al.}]{mnih2015human}
Mnih, V.; Kavukcuoglu, K.; Silver, D.; Rusu, A.~A.; Veness, J.; Bellemare, M.~G.; Graves, A.; Riedmiller, M.; Fidjeland, A.~K.; Ostrovski, G.; et~al. 2015.
\newblock Human-level control through deep reinforcement learning.
\newblock \emph{nature}, 518(7540): 529--533.

\bibitem[{M{\"o}hlenkamp and Thiele(2020)}]{mohlenkamp2020ventilation}
M{\"o}hlenkamp, S.; and Thiele, H. 2020.
\newblock Ventilation of COVID-19 patients in intensive care units.
\newblock \emph{Herz}, 45(4): 329--331.

\bibitem[{Peine et~al.(2021)Peine, Hallawa, Bickenbach, Dartmann, Fazlic, Schmeink, Ascheid, Thiemermann, Schuppert, Kindle et~al.}]{peine2021development}
Peine, A.; Hallawa, A.; Bickenbach, J.; Dartmann, G.; Fazlic, L.~B.; Schmeink, A.; Ascheid, G.; Thiemermann, C.; Schuppert, A.; Kindle, R.; et~al. 2021.
\newblock Development and validation of a reinforcement learning algorithm to dynamically optimize mechanical ventilation in critical care.
\newblock \emph{NPJ digital medicine}, 4(1): 32.

\bibitem[{Pham, Brochard, and Slutsky(2017)}]{pham2017mechanical}
Pham, T.; Brochard, L.~J.; and Slutsky, A.~S. 2017.
\newblock Mechanical ventilation: state of the art.
\newblock In \emph{Mayo Clinic Proceedings}, volume~92, 1382--1400. Elsevier.

\bibitem[{Shafer and Vovk(2008)}]{shafer2008tutorial}
Shafer, G.; and Vovk, V. 2008.
\newblock A tutorial on conformal prediction.
\newblock \emph{Journal of Machine Learning Research}, 9(3).

\bibitem[{Silva, Rocco, and Pelosi(2022)}]{silva2022personalized}
Silva, P.; Rocco, P.; and Pelosi, P. 2022.
\newblock Personalized mechanical ventilation settings: Slower is better!
\newblock In \emph{Annual Update in Intensive Care and Emergency Medicine 2022}, 113--127. Springer.

\bibitem[{Slutsky and Ranieri(2013)}]{slutsky2013ventilator}
Slutsky, A.~S.; and Ranieri, V.~M. 2013.
\newblock Ventilator-induced lung injury.
\newblock \emph{New England Journal of Medicine}, 369(22): 2126--2136.

\bibitem[{Suo et~al.(2021)Suo, Agarwal, Xia, Chen, Ghai, Yu, Gradu, Singh, Zhang, Minasyan et~al.}]{suo2021machine}
Suo, D.; Agarwal, N.; Xia, W.; Chen, X.; Ghai, U.; Yu, A.; Gradu, P.; Singh, K.; Zhang, C.; Minasyan, E.; et~al. 2021.
\newblock Machine learning for mechanical ventilation control.
\newblock \emph{arXiv preprint arXiv:2102.06779}.

\bibitem[{Sutton and Barto(1998)}]{sutton1998reinforcement}
Sutton, R.~S.; and Barto, A.~G. 1998.
\newblock Reinforcement learning: an introduction MIT Press.
\newblock \emph{Cambridge, MA}, 22447: 10.

\bibitem[{Van~Hasselt, Guez, and Silver(2016)}]{van2016deep}
Van~Hasselt, H.; Guez, A.; and Silver, D. 2016.
\newblock Deep reinforcement learning with double q-learning.
\newblock In \emph{Proceedings of the AAAI conference on artificial intelligence}, volume~30.

\bibitem[{Vazquez and Facelli(2022)}]{vazquez2022conformal}
Vazquez, J.; and Facelli, J.~C. 2022.
\newblock Conformal prediction in clinical medical sciences.
\newblock \emph{Journal of Healthcare Informatics Research}, 6(3): 241--252.

\bibitem[{Watkins and Dayan(1992)}]{watkins1992q}
Watkins, C.~J.; and Dayan, P. 1992.
\newblock Q-learning.
\newblock \emph{Machine learning}, 8: 279--292.

\bibitem[{Yu et~al.(2021)Yu, Liu, Nemati, and Yin}]{yu2021reinforcement}
Yu, C.; Liu, J.; Nemati, S.; and Yin, G. 2021.
\newblock Reinforcement learning in healthcare: A survey.
\newblock \emph{ACM Computing Surveys (CSUR)}, 55(1): 1--36.

\bibitem[{Yuan et~al.(2023)Yuan, Shi, Yang, Li, Cai, and Tang}]{yuan2023conservative}
Yuan, Y.; Shi, J.; Yang, J.; Li, C.; Cai, Y.; and Tang, B. 2023.
\newblock Conservative Q-Learning for Mechanical Ventilation Treatment Using Diagnose Transformer-Encoder.
\newblock In \emph{2023 IEEE International Conference on Bioinformatics and Biomedicine (BIBM)}, 2346--2351. IEEE.

\bibitem[{Zein et~al.(2016)Zein, Baratloo, Negida, and Safari}]{zein2016ventilator}
Zein, H.; Baratloo, A.; Negida, A.; and Safari, S. 2016.
\newblock Ventilator weaning and spontaneous breathing trials; an educational review.
\newblock \emph{Emergency}, 4(2): 65.

\end{thebibliography}
\endgroup

\end{document}